# ANUSAARAKA: OVERCOMING THE LANGUAGE BARRIER IN INDIA

Akshar Bharati, Vineet Chaitanya, Amba P. Kulkarni, Rajeev Sangal
IIT Kanpur Centre for NLP at Hyderabad
c/o CALTS, University of Hyderabad Campus

G. Umamaheshwar Rao
Centre for Applied Linguistics and Translation Studies
University of Hyderabad
{vineet,amba,sangal,guraosh}@uohyd.ernet.in

**ABSTRACT**

The anusaaraka system makes text in one Indian language accessible in another Indian language. In the anusaaraka approach, the load is so divided between man and computer that the language load is taken by the machine, and the interpretation of the text is left to the man. The machine presents an image of the source text in a language close to the target language.In the image, some constructions of the source language (which do not have equivalents) spill over to the output. Some special notation is also devised. The user after some training learns to read and understand the output. Because the Indian languages are close, the learning time of the output language is short, and is expected to be around 2 weeks.

The output can also be post-edited by a trained user to make it grammatically correct in the target language. Style can also be changed, if necessary. Thus, in this scenario, it can function as a human assisted translation system.

Currently, anusaarakas are being built from Telugu, Kannada, Marathi, Bengali and Punjabi to Hindi. They can be built for all Indian languages in the near future. Everybody must pitch in to build such systems connecting all Indian languages, using the free software model.

## 1. INTRODUCTION

Fully-automatic general purpose high quality machine translation systems (FGH-MT) are extremely difficult to build. In fact, there is no system in the world for any pair of languages which qualifies to be called FGH-MT. The reasons are not far to seek. Translation is a creative process which involves interpretation of the given text bythe translator. Translation would also vary depending on the audience and the purpose for which it is meant. This would explain the difficulty of building a machine translation system. Since at present, the machine is not capable of interpreting a general text with sufficient accuracy automatically - let alone re-expressing it for a given audience, it fails to perform as FGH-MT. FOOTNOTE{The major difficulty that the machine faces in interpreting a given text is the lack of general world knowledge or common sense knowledge, subject specific knowledge, knowledge of the context, etc. which can collectively be called as background knowledge.}

The first difficulty that the machine faces, occurs at a level which we normally do not even recognize as a problem. It pertains to information coded in a text.

To understand the idea of information and its coding, let us consider an example. In Indian languages, which have relatively free word-order, information that relates an action (verb) to its participants (nouns) is primarily expressed by means of post-positions or case endings of nouns (collectively called vibhaktis of the noun). FOOTNOTE{Such a relation between the action and its participant is called a karaka relation.} For example, in the following sentence:

rAma ne roTI khAI (1)
Ram-erg. bread ate
Ram ate the bread.

The ergative (erg.) post postion marker ('ne') after 'rAma' indicates that Ram is the *karta* of eat, which here means that Ram is the *agent* of eating. (Note that in English, the primary device for expressing the same information is by means of word order.)

Noun-verb agreement also helps in identifying the karta. For example, in the following sentence:

rAma roTI khAtA hE (2)
Ram (m.) bread (f.) eats (mase)
Ram eats bread.

the masculine ending of the verb indicates that the karta is masculine, which in this sentence unambiguously means Ram. However, this is not always unambiguous; consider the following sentence:

chAvala rAma khAtA hE (3)
rice (m.) Ram (m.) eats (m.)
Ram eats rice.

in which the agreement does not help in identifying the karta unambiguously. There are two masculine nouns (Ram or rice) one of which is the karta. Translation to English, say, would be quite different depending on which one is the karta.

This example raises an important point. A language text actually "codes" or contains only partial information. When a reader (or listener) interprets the text by suitably supplying the missing information, he get the intended meaning. A text is akin to a picture made up of strokes (and gaps). A viewer fills in the missing parts appropriately and views them as part of the picture. If done properly, the reader gets the message intended by the writer. (In language, there is a tension between brevity and ambiguity. If everything was explicity stated, the text would be less ambiguous but would be long. Brevity also helps in focussing attention to the relevant parts. Ambiguity seems to be a necessary price for conciseness and focus.)

## 2. WHAT THE MACHINE CAN AND CANNOT DO

### 2.1 WEAKNESS

To understand the nature of difficulty the machine faces, let us return to the example sentence (3). It seems trivial for us to assume that a person (Ram, in this sentence) would be the agent of eating, and rice, the thing which is eaten. But the machine does not "know" that. This knowledge is said to be world knowledge, as it pertains to the world as it exists. It turns out, that if we try to put this kind of information in the mchine, we find that there are a very large number of such facts. For all the nouns, we will have to say who can eat whom. How should such facts be organized, is the first problem? But there is a still harder problem that turns up. Such knowledge is quite easily over-ridden in language to convey a metaphorical sense, irony, etc Consider the following sentence:

SarAba Apa nahiM pIte, SarAba Apako pItI hE (4)
Alcohol you not drink, alcohol you (accus.) drinks.
You do not drink alcohol, alcohol drinks you.

Here, the sentence says that the alcohol drinks a person! And it is a perfectly good sentence.

Thus, it is not enough to put such a large number of facts about the world in the machine, we must also put conditions regarding when they can be overridden while processing text. This turns out to be an incredibily hard task. This is the major problem, which the discipline of Artificial Intelligence is addressing, but with only limited success. There are no known methods by which the machine can handle and use world knowledge satisfactorily today, while processing unrestricted language text.

The examples we have considered, are rather easy because the world knowledge that needs to be referred to is fairly well pin-pointed. Quite frequently, there is ellipses in sentences (i.e. parts of sentences are dropped). The missing part may have to be inferred (possibly using world knowledge) before the processing can be done further.

## 2.2 STRENGTH

We have just seen a major weakness of the machine. It has little or no common sense or world knowledge. Therefore, it cannot interpret or use judgment well. But there are two aspects in which it is strong. It has a large memory, and it can perform arithmetic and logical operations very fast. For example, it can easily store a large dictionary of a few hundred thousand words, and it can search a given word very quickly. Similarly, if the machine is given a grammar rule, it can apply it faithfully and with great speed. Much of language related data and rules can be fed into the machine, more easily than background knowledge.

# 3. SHARING THE LOAD

Does it mean that since the machine cannot interpret text with a fair degree of accuracy, machine translation must be abandoned as a distant dream? The answer lies in sharing the load between the reader and the machine so that the tasks which are hard for the human being are done by the machine and vice versa.

A clean way to share the load is for the machine to take up the task of language related processing, and to leave the processing related to background knowledge to the reader. Language related processing consists of analysis of the input source language text such as morphological processing, use of bilingual dictionary, and any other language related analysis or generation. These are the primary sources of difficulty to the reader. These are also the tasks which are relatively easier for the machine. On the other hand, world knowledge related aspects are left to the reader, who is naturally adapt at it.

In translation, two opposing forces are at work: faithfulness and naturalness. The translator must chose between faithfulness to the original text and naturalness to the reader. If the translation is to be made easy and natural to the reader, the translator may have to depart from the original text, and put it in a style and setting familiar to the reader. But then the flavour and subtlety of the original gets lost. For a reader, who wants to read and study what the original writer wrote, such a translation is not satisfactory. This also means that there is no unique "correct" translation - in fact, the appropriateness of a translation depends on the audience and the purpose it is meant for.

Most translations that we come across, are weighted towards naturalness to the reader. Anusaaraka is at the other extreme; it tries to be as faithful to the original text as possible. In fact, its output must contain all the information in the source language text, and no other new information. FOOTNOTE{It can, however, present the information in stages. Discussed later.}

There is a problem in coding "exactly" the same information (with 100% fidelity) from one language to another, particularly if we want to generate sentences of about equal length, paralleling the sentence constructions wherever possible. (In this sense, translation is sometimes said to be an impossible task). FOOTNOTE{This also suggests the incommensurability of information. When some information is transferred from one language to another, there is no way to express it exactly. If one tries to move from one construction to another some part of the information is missed, per force. To take a mathematics example, there is no way to express the cube root of 3 in rational numbers, real numbers are needed. However, it can be approximated to a given degree of precision.}

The anusaaraka answer lies in deviating from the target language in a systematic manner whenever necessary. This new language is something like a dialect of the target language. For example, the Kannada to Hindi anusaaraka is likely to produce the following Hindi from a normal Kannada text:

@H: mohana kala AyegA EsA rAma kahA (5)
!E: Mohan tomorrow will-come this Ram said

because Kannada uses this kind of construction. The normal usage in Hindi:

H: rAma ne kahA ki mohana kala AyegA. (6)
!E: Ram-erg. said that Mohan tomorrow will-come

The anusaaraka output can be said to be the image of the source text, much like what the camera produces. Reading the image of source text is like reading the original text. It will have the same flavour. Translation, on the other hand, is like a painting. The translator interprets the original in the source language, and "paints" a text in the target language with the same meaning.

Readers will usually require some learning of the dialect of the target language (discussed in Section 6). This learning time will be negligible compared to the learning time of the source language.

## 4. SOME STANDARD COMPONENTS OF AN MT SYSTEM

In this section we will take a look at some of the components to perform some standard tasks in any machine translation system. We will also discuss how they can be put together. In Section 5, we will return to a discussion of issues regarding relatedness of Indian languages, language bridges, etc., and how anusaaraka makes use of them.

A basic MT System consists of an analyzer of the source language whose output is fed into the generator of the target language. Between the analyzer and the generator there is a mapper which uses bilingual dictionaries to map the source language elements to target language elements. The important components are described below.

## 4.1 WORD ANALYSER

Words in the input text are first processed by the morphological analyzer. Its task is to identify the root, lexical category, and other features of the given word. For example, for the Telugu word 'mAnava', its morphological analysis yields two possibilities: noun and verb.

1. mAnavuDu{category=noun,number=sg,case=oblique}
   The above Telugu root 'mAnavuDu' means: mAnava or man

2. mAnuvu{cat=verb,TAM=infinitive,gnp=any}
   The above Telugu root 'mAnuvu' means: ghAva_bharanA or heal

   (gnp stands for gender-number-person, TAM for tense-aspect-modality.)

In the case of noun, number and case are shown, and in the case of the verb, TAM label and gnp are shown. Some more examples are given below.

```
smRti      smRti{cat=n,number=sg,case=0}/
           smRti{cat=n,number=sg,case=oblique}
vyAdhulaku vyAdhi{cat=n,number=pl,case=ki}
agunaTlu   avvu{cat=v,TAM=jEsA,gnp=any}
jAta       jAta{cat=n,*adj_0* }/jAta_adj_n{n sg *obl* }/
           jAta_adj_m{n sg *obl* }
telipiri   telupu{cat=v,TAM=*iti*,gnp=non-neuter_pl_3}
```

The morphological analyzer we describe is designed to handle inflectional morphology. (Separate module would be needed for derivational morphology.) For a given word, it checks whether the word is in the dictionary. If found, it returns its lexical category (such as pronoun, post-position, noun, verb, etc.) and other grammatical features. It also tries to see whether the word can be broken up into a root and a suffix. At the breakup point, some characters such as vowels may be added or deleted. It may have to try several times proposing to break the word at different points. For each proposed breakup, it looks up the proposed root in a dictionary and the proposed suffix in a suffix table. Whenever, both lookups are successful, it is a valid root and suffix. FOOTNOTE{Provided they are compatible to each other, information about which is also stored in the dictionary.} From these, information is returned regarding the root, its lexical category and the grammatical features.

If the above morphological analysis does not yield any answer, compounding or sandhi breaking is tried. The given word is broken up into two parts, and each part is analyzed as a proposed word. (Thus, for each of the two parts, the morphological analysis outlined above is repeated, which might again result in proposing roots and suffixes etc. for each proposed word.) This method is called propose and test method.

A large number of steps may have to be tried in the above procedure. There are ways of speeding up or eliminatng some of the steps. But since each step is mechanical and small the machine can carry it out precisely and fast.

## 4.2. LOCAL WORD GROUPER

Indian languages have relatively free word-order, still there are units which occur in fixed order. In Hindi, the most important examples of these are the nouns followed by post-positions, main verb followed by auxiliaries, or compound nouns. In general, whenever there are a sequence of words that have a meaning which cannot be composed out of the meanings of individual words, they must be grouped together and the group as a whole will have a meaning. The group as a whole together with its meaning will have to be stored in a dictionary or a table. Some examples are given below. (Labels H and E specify the language of the sentence as Hindi and English, respectively, and '!E' stands for gloss in English.)

H: laDake ke lie
!E: boy for
E: for the boy

H: khAta calA jA raHA HE
!E: eat walk go live is
E: going on eating (without stopping)

H: kAlA pAnI
!E: black water
E: rigorous imprisonment.

Local word grouping is more extensive in Hindi and other north Indian languages compared to the south Indian languages, while morphology is simpler. Thus, the two taken together (morphology and local word grouping), are likely to have the same level of difficulty across the north and south Indian languages.

## 4.3. MAPPER USING BILINGUAL DICTIONARIES

This process involves looking up the elements of the source language and substituting them by equivalent elements belonging to the target language. For example, the root of a source language word obtained using a word analyzer is substituted by equivalent root in the target language. For example, 'Apa' would be produced in Hindi for the Telugu word 'mIru' (you). The grammatical features also need to be mapped suitably. For example, a pronoun and a noun in the source language (mIru and pustakaM) respectively are mapped to an appropriate pronoun and noun, in the target language below with the same number, person, etc.

T:  mIru  pustakaM  caduvutunnArA?                    (7)
@H:  Apa   pustaka   paDha_raHA_[HE|thA]_kyA{23_ba.}?
!E:  You   book      read_ing_[is|was]_Q.?
E:  Are/were you reading a book?

(Where the labels mean the following:
T=Telugu, @H=anusaaraka Hindi, !E=English gloss, E=English.)

In the example above, the last word in the sentence is a verb and illustrates the mapping from Telugu to Hindi, morpheme by morpheme: the root is mapped to 'paDha' (read), and similarly the tense-aspect-modality (TAM) label is mapped to 'raHA_[HE|thA]' (is_*ing or was_*ing), which is followed by 'A' suffix which gets mapped to 'kyA' (what) as a question marker in Hindi. Telugu leaves the tense open as: present or past, which is reflected in the output. GNP information is also shown separately in curly brackets ('{23_ba.}' for second or third person and bahu-vachana (plural)).

### 4.4 WORD SYNTHESIZER

A word synthesizer is the reverse of the word analyzer. It takes a root, its lexical category and grammatical features, and generates a word. Two examples in Hindi are given below:

rAjA{cat=noun,number=pl,case=oblique}        ===> rAjAoM
king

khA{cat=verb,number=sg,TAM=tA,gnp=fem_sg_3}  ===> khAtI
eat

Word synthesis is a much simpler task compared to word analysis.This can usually be done directly by the given rules, without having to try various alternatives, by proposing and testing.

### 4.5 PUTTING THE COMPONENTS TOGETHER

The above components can be put together, resulting in an MT system. A sample system is described below, but there can be variations on this theme.

First, the input text in a source language is passed through its word analyzer, which analyzes each word and produces its root and grammatical features. These are fed into a local word grouper, which combines the words and produces local word groups. Second, the mapper takes the output produced so far, to replace the elements of the source language with elements of the target language. Thus, at this stage, the source language root will be changed to target language root. Third, the output of the mapper is fed into the generator of the target language, which itself might consist of local word spliter and morphological synthesizer. The output produced is the MT system output.

Interfaces are also provided for human pre-editing of the input and post-editing of the output. These are also a part of the overall system, but a discussion on them is postponed to Section 6.

## 5. ANUSAARAKA APPROACH

As explained earlier, anusaaraka takes the information in the source language text and presents it in the target language (or in a language close it). Thus, at the suffix level, a suffix in the source language is replaced by a suitable element in the target language; and at the word level, the source words are replaced by equivalent words in the target language. Similarly, the word groups are also replaced by equivalent groups, etc. in the target language. The reason the above approach works even without a parser, is that Indian languages are syntactically similar.

Indian languages are relatively free word-order where the noun-groups can come in any order followed generally by the verb group. (The order conveys emphasis etc. but not the information about karaka relationships.) If we take a sentence in a source language, and substitute the word groups in it by appropriate word groups in the target language, it works well because the languages make similar use of order to convey emphasis etc. The vibhaktis for the word groups (that is, case endings and post-position markers for nouns, and TAM for the verb groups), must be mapped from the source language to the target language carefully, as they contain important karaka information regarding the verb and the nouns. Again the languages behave in a similar way.

Besides the above, there are similarities in the meanings of words. Many words in the languages have a shared origin (from Sanskrit), and because of shared culture, they usually also share meanings. This implies that for a source language word, the bilingual dictionary provides a unique answer in the target language.

Thus, the reason why the method outlined in the last section works well is the above similarity among Indian languages. Even if the languages have a different origin, if they are in close contact, they acquire each others' features. This is called the 'areal hypothesis'. Scholars further agree on calling India as a linguistic area.

Now, we will discuss some problems becaause the two languages differ, and see how these problems can be handled. We will take examples from Hindi, Telugu and Kannada. Apart from agreement, there are only three major syntactic differences between Hindi and Kannada. Surprisingly all of these can be taken care of by enriching Hindi with a few additional functional particles or suffixes as shown below. Thus, they can be viewed as lexical gaps or function word gaps. But first we will discuss issues related to agreement.

## 5.1 AGREEMENT

Let us consider the case of noun-verb agreement. There is a lack of agreement (of gender, number and person) as per the rules of the target language in the anusaaraka output. The information about gender etc. is displayed corresponding to the source language. For example, in the anusaaraka output below, the masculine and feminine gender is marked by {m.} and {f.} resepectively against the personal pronoun 'vaHa'. ({~m.} stands for non-masculine). Note that in Hindi, personal pronoun 'vaHa' is the same for both masculine and feminine gender.

T: Ame vADito mATlADiMdi kAnI, (8)
@H: vaHa{f.} usa{m.}_se` bAta_kiyA_[HE|thA]{3_~m._e.} lekina[Hone_do],
!E: she he(instr.) talked(non-masc.) but, (9)

T: vADu Ameto mATlADaledu.
@H: vaHa{m.} usa{f.}_se` bAta_kiyA_naHIM[naHIM_bAta_kara_sakatA_HE{3_~m._e.}].
!E: he she(instr.) did[could]_not_talk(non-masc.).

E: She talked to him, but he did not talk to her.

If the gender information was not shown, the sentence rendered would have been rendered as:

H: usane usase bAta kI, lekina usane usase bAta nahIM kI. (10)
!E: s/he s/he(instr.) talked, but s/he s/he(instr.) talk not do

Without the gender information in anusaaraka Hindi, the meaning of the sentence is not clear. To produce good Hindi from such a sentence, requires different strategies. One solution would be to explicitly add 'laDakA' (boy) etc. indicating the sex:

H: usa laDakI ne usase bAta kI, lekina usa laDake ne (11)
!E: that girl-erg. her/him talked, but that boy erg.

H: usase bAta nahIM kI.
!E: her/him talked not

But whether it should be boy, or a man or something else would depend on the context, and quite beyond the capability of the machine to infer correctly in all possible situations. Another solution would be to change the tense-aspect label slightly, so that it becomes different from past-completive (at the cost of faithfulness to the orginal). By doing this, karta-verb agreement would no longer be blocked by the post-position marker, and show the gender in the verb. Yet another solution would be to use 'bolatA hE' (speaking) a construction in which agreement between noun-verb specifies the gender of the karta of the speaker.

H: vaha usase bolI, para vaha usase nahIM bolA. (12)
!E: spoke(f.) spoke(m.)

Appropriate selection and use of such strategies is left to the post-editor in the anusaaraka approach. (Post-editor is a reader who is editing the output to make it grammatically correct and suitable for wider use. This issue is discussed in Section 6.2. Some interfaces are provided so that such a user can make changes with ease.)

## 5.2 "ki" CONSTRUCTION

In case of embedded sentences in Hindi, the subordinate sentence is put after the main verb unlike in Kannada. For example:

H: rAma ne kahA ki mEM ghara ko jAUMgA. (13)
!E: Ram erg. said that I home acc. will_go
E: (Ram said that he will go home.)

There is a construction in Kannada which is similar (below, label 'K' stands for Kannada):

K: rAma heLidanu eneMdare nAnu manege hoguttene.
@H: rAma kahA ki mEM ghara_ko jAUMgA.
!E: Ram said that I home_acc. will_go
E: (Ram said that he will go home.)

However, it is seldom used. Kannada uses another construction for which the anusaaraka Hindi is given below (repeated from sentence (5)).

K: mohana nALe baruvanu eMdu rAma heLidanu. (5)
@H: mohana kala AyegA EsA rAma kahA.
!E: Mohana tomorrow come-fut that Rama said.

'EsA' construction is a proper construction in Hindi; only it is used less frequently. In the dialect of Hindi produced by anusaraka from south Indian languages however, this will be the normal construction used.

## 5.3 "jo" CONSTRUCTION

In this section, we will discuss how anusaaraka handles participle verbs (behaving as adjectives) in Kannada to produce the same information in Hindi. The solution works for all south Indian languages, which display this phenomenon.

We will first try to derive the meaning of TAM labels which stand for adjectival participle, in a mathematically precise way. Let us take the following Telugu example sentence:

```
  T: rAmuDu tinina camacA veVMDidi.              (14)
     ------ --- -- ------ --------
       1    2a  2b    3      4
 !E: Ram   *eaten  spoon  silver-of
  E: The spoon with which Ram ate is of silver.

  (* 'eaten' is only an approximation, 'tinina' is a
   past-participle form of 'tina' or 'eat')
```

We are interested in finding the meaning of the TAM label or suffix 'ina' suffix in 'tinina' above. Let us name it 2b, and the rest of the words are also named for easy reference.

If a Telugu-Hindi bilingual person is asked to translate the sentence, he is likely to write down the following in Hindi:

```
  H: rAma ne  jisa  cammaca se   khAyA, vaHa cAMdI kA HE.
     ---- ++        -------      ---         -------- ++
      1             3            2a            4
 !E: Ram erg. which spoon instr. ate,   that silver_of is
```

Here the Hindi words are marked corresponding to the Telugu words (other than 2b whose value we want to find out). '++' is used to denote words that have been put by the translator but which are not there in the original Telugu sentence. 'ne' corresponds to the ergative marker which is an idiosyncracy of Hindi. Also it is known that 'HE' at the end (copula) is mandatory in the Hindi sentence but is absent in the given Telugu sentence.

We can rephrase the sentence in Hindi to get the words in the same order

```
  H: rAma ne jisa se khAyA HE vaHa cammaca cAMdI kA HE.
     ---- ++         ---          ------- -------- ++
      1              2a             3        4
```

or better still, we may rewrite the above as:

```
  H: rAma ne khAyA  HE  jisa  se     vaHa cammaca cAMdI kA HE.   (15)
     ---- ++ ---                          ------- -------- ++
       1     2a                             3       4
 !E: Ram erg. eaten has which instr. that spoon   silver_of is
```

wherein the order of the words including the parts of words (2a and 2b) is exactly the ame as the order in the original sentence. Now the part which remains unassigned, stands for 2b. Therefore, we get the equation:

ina = yA_HE_jisa_se_vaHa

But a closer scrutiny reveals an assumption, "se" or instrumental marker is not there in the Telugu sentence. For example, consider the following sentence:

```
  T: rAmuDu winina pleTu veVMdixi                    (16)
     ------ --- -- ------ --------
       1    2a  2b   3      4
 !E: Ram    eaten  plate  silver-of
  E: The plate in which Ram ate is of silver.
```

Its equivalent Hindi sentence is:

```
  H: rAma ne khAyA HE jisa meM vaHa pleTa cAMdI kI HE.     (17)
     ---- ++ ---                  ----- -------- ++
       1     2a                     3      4
```

The above sentence yields the following equality:

ina = yA_HE_jisa_meM_vaHa

The two different equalities for 'ina', and similar other examples lead us to conclude that the 'se' or 'meM' markers are not there in the 'ina' but are supplied by the reader based on the world knowledge. Therefore, the equality becomes:

ina = yA_HE_jo_*_vaHa

where '*' stands for an unspecified post-position to be supplied based on context. The claim is that the above is a mathematically precise equivalence between the 'ina' Telugu TAM and anusaaraka Hindi.

The above can be restated as follows: It shows the equivalence between the adjectival participle in Telugu and the relative clause in Hindi, which has been known, but which the above equation makes precise. Although, Hindi also has participial phrases it has only two TAMS: yA and tA_HuA (with perfective and continuous aspects, respectively).

```
 H: khAyA HuA phala        (18)
    eaten     fruit

 H: khAtA HuA hiraNa       (19)
    eating    deer
```

As a result, these are not sufficient to capture other TAMs which might occur in Telugu. There is a gap in Hindi.

There is another problem, too, as we have seen. The two participal phrases in Hindi have coding for karaka relations which is absent in Telugu. TAM 'tA_HuA' codes karta karaka (roughly agent), and the sentence 5.8 says, the deer who is eating (not the one who is being eaten). Similarly, yA codes karma as in sentence 5.7 (the fruit being eaten, and not the fruit who is eating). FOOTNOTE{More correctly, yA codes karma in case of sakarmaka or transitive verbs, and karta in case of intransitive verbs.} Thus, Hindi is poorer than Telugu in coding tense, aspect, modality information, while richer in coding karaka information. But this creates another difficulty for anusaaraka. Using these constructions in Hindi, would mean putting in something that is not contained in the source language sentence, and the information equivalence would be lost.

To take care of the limitation of the TAMs, we select relative clause construction in Hindi. This, however, also requires the karaka information to be specified. To express the same information as in the Telugu sentences (5.3) and (5.5), we have invented a notation along with the jo-construction as described earlier.

'jo_*_vaHa' could even be replaced by 'so' to produce a kind of colloquial Hindi in south India (Dakkhini Hindi).

khAyA HE so cammaca

Unlike the 'ki' construction (Section 5.2), this idea takes some time and effort for the Hindi reader to get used to.

### 5.4 "ne" CONSTRUCTION

The "ne" construction or ergative marker is a peculiarity of only the Western belt languages in India. In case of the present or past perfective aspect of the main verb in Hindi sentence, "ne" is used with the karta:

H: rAma ne phala khAyA. (20)
!E: Ram erg. fruit ate.
(Ram ate the fruit.)

In anusaaraka output from Kannada to Hindi, the 'ne' post-position would never be produced. It would not be produced even with the TAM label 'yA' in Hindi (wherein it is mandatory barring a few exceptions verbs). For example:

H: rAma phala khAyA. (20')

Therefore, we can postulate a new TAM (yA`) with same semantics as "yA", but which does not use "ne" construction in anusaaraka Hindi. With this TAM, we can express the corresponding Kannada sentence more faithfully as:

It may be of interest to note that the "yA" pratyaya in Hindi corresponds to "kta" pratyaya in Panini's grammar and so the new proposed pratyaya (yA`) will be a natural counterpart of the "ktavatu" pratyaya in the Sanskrit grammar.

Thus, in this section we have tried to show how the differences among the languages are bridged and the information is carried across. The reader might need some training to read the anusaaraka output.

## 6. PRE-EDITING AND POST-EDITING

Anusaaraka system has been designed so that the combination of man and machine together can perform translations, etc. We have earlier said that the tasks which are routine can be handed over to the machine, and those difficult for the machine are left for the user. In this section, we will briefly discuss the user intervention in the task. FOOTNOTE{We have said that all the information in the source text is preserved in the output. Although it preserves information in the output, it can present the information in stages. The raw output, which is the first output that a user sees, might not show all the details. Only when the user requests for the details, they are shown. The raw output could also be tuned to the requirements of a user, and thus could be made different for different class of users.}

There are two principal points in this whole process at which the user can help: pre-editing the input and post-editing the output.

### 6.1. PRE-EDITING AND LANGUAGE VARIATION

In the pre-editing task, the input text is corrected and edited by the user: Words spelt with non-standarad spellings are changed to their standard spellings, external sandhi between words is broken (unless it changes meaning), etc.

This is an important task for Indian Languages because of lack of standardization and consequent variation. It is particularly serious in Telugu. Spelling variation is very large. On an average a word can be written in three alternate ways. Partly the reason for this is that the written material has been influenced by the local dialects in the last forty years. In fact, use of local dialects of Telugu in written texts was actively promoted by the young and influential writers in this period. There has also been no major effort at standardization.

Similarly, there are lack of standards in the use of space. Sometimes sandhi between words is performed, sometimes not so. Worse still for the machine, when sandhi results in a long word, it is broken up at a point different from where the sandhi has been done. The machine will thus have difficulty with both the resulting words.

Speling variation might be severe in written Telugu, but is present in all Indian languages. Much more so than say in English. One will have to live with this reality, while designing MT systems.

It might be argued by some that machine must handle all the variations. For the phenomena mentioned above, in principle, it does not seem to be a problem. However, in practice, it requiries a much bigger effort. Instead of three years it might mean twenty years to develop a working system. Therefore, for the machine to start doing something useful, it becomes important to handle a sub-language first, say, the standard language (to the extent defined already, or by extending the definition). However, the sub-language should be so chosen that sufficient amount of written material exists which is needed by other language groups or persons, for the system to be useful.

A pre-editing interface can help the human pre-editor. The pre-editor can run such an interface software, which points out the non-standard forms and seeks corrections It can also present alternatives out of which he can choose the correct form.

## 6.2 POST-EDITING THE ANUSAARAKA OUTPUT

It has already been discussed that the ansuaaraka output is close to the target language, and in general is not grammatical from the target language viewpoint. In case, a user is reading for his own sake, he might not bother to produce a grammatically correct and stylistically more suitable output. However, when a document is going to be distribued in large numbers, it would normally be post-edited by a person before distribution or publication.

There are three levels of post-editing. the first level of post-editing seeks to make the output grammatically correct. The emphasis is on speed and low cost. The posteditor might drop phrases, change construction in the interest of speed, as long as it does not alter the gross meaning. Under this level of post-editing, corrections are made regarding agreement, putting 'ne' (ergative marker) where necessary (see Section 5.4), inserting the correct vibhakti in jo_* construction (see Section 5.3), etc.

In the second level of post-editing the raw output is corrected not only grammatically but also stylistically. There can be many different types and quality of output at this level, depending on the audience. One audience might be willing to accept some constructions in the raw output which are grammatically correct in Hindi but not used often. Another audience might not be willing to accept it . For example, 'EsA' construction (see Section 5.2) can be changed to 'ki' construction, for such an audience.

In the third level of post-editing the post-editor might change the setting and the events in the story to convey the same meaning to the reader who has a different cultural and social milieu. This is really trans-creation, and a creative post-editor can go all the way upto this level.

A post-editing interface allows him to do post-editing rapidly. Rather than making corrections character by character, he can supply the missing information and the computer can carry out the corrections. For example, to make a verb form into feminine plural, he need not individually change the verb and its auxiliaries to the correct forms manually. Instead, he can place the cursor on the verb sequence and give a command, and the computer would change the forms of the verb and its auxiliaries.

## 6.3 TRAINING

The reader of the Anusaaraka output would need to undergo training. Besides covering the special symbols used in the output, the training would also familiarize him with the differences in the source and target language. This is important because he is likely to encounter constructions of the source language in the output, the output being an image of the source text. Where the constructions in the two languages are similar, the output will be transparent, but when they are different, he would need to know the construction in the source language. It is hoped that such a training will take about two weeks.

There would be additional training for post-editors. It would teach them about the different levels of post-editing and how to choose among them based on requirements. It would also familiarize them with the computer interface which speeds up post-editing.

# 7. CONCLUSION

## 7.1. SUMMARY

We have discussed the anusaaraka approach to building computer software so that text in one Indian language becomes available in another Indian language. In this approach, the load is so divided between man and machine that the language load is taken by the machine, and the interpretation of the text is left to the man. The machine presents an image of the source text in a language close to the target language. The user after some training learns to read and understand the text in this language. The output can also be post-edited by a trained user to make it grammatically correct. Style can also be changed, if necessary.

## 7.2. STATUS

Anusaarakas can be built for all Indian languages in the near future. Currently, anusaarakas are being built from Telugu, Kannada, Marathi, Bengali and Punjabi to Hindi. They can also be built in the reverse direction. In fact, it is useful to group the anusaarakas for the different languages:-

(a) South Indian languages to Hindi and vice-versa.
(Tamil, Telugu, Kannada, Malayalam, etc.)

(b) Eastern languages (Bengali, Assamese, Oriya, etc.)

(c) Western languages (Konkani, Marathi, Gujarati, etc.)

(d) Northern languages (Punjabi, Kashmiri, Urdu, etc)

There will be many similarities among the anusaarakas within the same group. Effort also needs to be made to build such systems among languages within the same group, for example, from Telugu to Kannada. This task is easier.

Finally, we address the issue of connecting to English. The task of building an anusaaraka system between English and Indian languages is harder, because English has a very different structure and vocabulary. It will take some time (probabaly around 5 years) for such a system to be built with enough power that it can be used effectively. Even then the system might require a longer training than the systems between Indian languages, before it can be used. However, even if one such system is built, it will make material from English available in one Indian language. The material can then become available in all other Indian languages through the other anusaarakas.

The anusaaraka output requires some effort and training to understand. For narrow subject areas, specialized modules can be built which produce good quality grammatical output. However, it should be remembered, that such modules will work only in narrow areas, and will sometimes go wrong. In such a situation, anusaaraka output will still remain useful.

## 7.3 IMPLICATIONS

If the anusaarakas enter into common use, it has major implications for national integration. The users of anusaaraka through both training as well as exposure to the raw output, will learn the features of the source languages they read. Thus, a reader of anusaaraka Hindi will learn features of the South Indian language if he uses a Telugu to Hindi anusaaraka. Many new constructions will also enter into the language. For example, the Hindi readers willsee constructions of say, Kannada, Bengali, Marathi etc. if they use anusaarakas. This will serve to broaden the target language. Thus, on the one hand while it will encourage people to work in their own languages, and thus strengthen the various Indian languages; on the other hand, it will further contribute to the mixing of languages, through a natural process of use (of viewing of images of other languages through the anusaaraka).

This has implications for the three-language formula too. As part of this formula, if anusaaraka is taught in schools and training is provided for it (which includes a study of important differences among Indian languages), the child would learn to access written text in not just one additional language but all Indian languages. Thus, written material including literataure, magazines, newspapers, official documents, become accessible to the person. FOOTNOTE{It is expected that in the near future (within a few years), computer-networks will spread and many of such texts will become available online. Libraries will also go online in a big way. At that time, such a training so that a person can access documents in any Indian language will be a big asset.}

When a child or a person, moves from one language region to another, he might have to learn the spoken form of the language as well. But acquiring the spoken form will be that much easier, because he will already be familiar with the constructions of the language.

For the above to become a reality, everybody must chip in to build dictionaries and prepare other language data, so that anusaarakas can be built connecting all Indian languages. Government can support this activity; but what is needed is for volunteers to come forward for the task. This should happen for the love of our languages. The right model to be followed for building the systems is the "free software" model, in which everybody contributes to the effort, and the results are open and available for everyone to use. People who contribute to building of the system, are acknowledged for their work, but nothing is hidden and nobody owns the software. This way one can begin where others have left off, and build on top of each others' work, in a cooperative activity. This can very well become a national people's cooperative project which can be emulated in other fields as well.

## REFERENCES

Natural Language Processing: A Paninian Perspective, Akshar Bharati, Vineet Chaitanya, Rajeev Sangal, Prentice-Hall of India, 1995 (Chap. 7).

Anusaraka: A Device to Overcome the Language Barrier, V.N. Narayana, Ph.D. thesis, Dept. of CSE, I.I.T. Kanpur, 1994.